\documentclass[11pt,a4paper]{article}
\usepackage{naacl2021}
\usepackage{times}
\usepackage{latexsym}
\usepackage{graphicx}
\usepackage[many]{tcolorbox}
\usepackage{url}

\title{The Spotify Podcast Dataset}

\author{Ann Clifton  \\
  {\normalsize \tt aclifton@spotify.com} 
  \And
 Aasish Pappu  \\
  {\normalsize \tt aasishp@spotify.com}\\
  \And
 Sravana Reddy  \\
  {\normalsize \tt sravana@spotify.com}
  \AND
  Yongze Yu  \\
  {\normalsize \tt yongzey@spotify.com}
  \And
 Jussi Karlgren  \\
  {\normalsize \tt jkarlgren@spotify.com}
  \And
Ben Carterette \\
  {\normalsize \tt benjaminc@spotify.com}
  \AND
  Rosie Jones \\
  {\normalsize \tt rjones@spotify.com}
  }

\date{}

\begin{document}

\maketitle
\begin{tcolorbox}[colback=red!10!white,
                     colframe=red!20!black,
                     title=\textsc{Updated Version Available},  
                     center, 
                     valign=top, 
                     halign=left,
                     before skip=0.8cm, 
                     after skip=1.2cm,
                     center title, 
                     width=3in]
 
     This paper is replaced by the COLING 2020 paper \cite{clifton-etal-2020-100000}, which elaborates on the podcast dataset, and evaluates benchmarks for retrieval and summarization. Please refer to the COLING paper instead of this document.
  \end{tcolorbox}

\begin{abstract}
    Podcasts are a relatively new form of audio media. Episodes appear on a regular cadence, and come in many different formats and levels of formality. They can be formal news journalism or conversational chat; fiction or non-fiction. They are rapidly growing in popularity and yet have been relatively little studied. 
As an audio format, podcasts are more varied in style and production types than, say, broadcast news, and contain many more genres than typically studied in video research. The medium is therefore a rich domain with many research avenues for the IR and NLP communities. 
We present the Spotify Podcast Dataset, a set of approximately 100K podcast episodes comprised of raw audio files along with accompanying ASR transcripts.
This represents over 47,000 hours of transcribed audio, and is an order of magnitude larger than previous speech-to-text corpora.
\end{abstract}
\section{Introduction}
Podcasts are a relatively new form of audio media. Episodes appear on a regular cadence, and come in many different formats and levels of formality. They can be formal news journalism or conversational chat; fiction or non-fiction. They are sharply growing in popularity and yet have been relatively little studied. 
As an audio format, podcasts are more varied in style and production types than, say, broadcast news, and contain many more genres than typically studied in video research. The medium is therefore a rich domain with many research avenues for the Information Retrieval and Natural Language Processing communities. 

% moved from related section: Podcasts are a new and emergent channel of spoken communication, and we can expect podcasts to settle into a set of genres as the produced material rises to meet use cases over time. 

In this work, we present the Spotify Podcasts Dataset, the first large scale corpus of podcast audio data with full transcripts. 
This corpus is drawn from a variety of heterogeneous creators, ranging from professional podcasters with high production values to amateurs without access to state-of-the-art production resources.
The podcasts cover a wide range of topics including lifestyle and culture, storytelling, sports and recreation, news, health, documentary, and commentary.
In addition, the content is delivered in a variety of structural formats, number of speakers, and levels of formality, whether scripted
or improvised, or in the form of narrative, conversation, or debate. 
These data present an interesting challenge for existing tasks such as spoken document retrieval, segmentation, and summarization, and will enable new avenues of speech and NLP research.

We release these data in collaboration with the TREC Text Retrieval Conference, along with a summarization and an ad-hoc retrieval challenge.
For more information about participating in the challenge and obtaining the dataset , please see the TREC website\footnote{\url{https://trec.nist.gov/pubs/call2020.html}}.

\section{Speech Datasets}

Existing collections of spoken material and speech-to-text corpora have characteristics that differ importantly from this dataset. 
As noted above, podcasts extend spoken language communication in certain ways and collections of previous material do not capture that entire range of variation. 

%Firstly, they are all an order of magnitude smaller than the podcast corpus, with X being size A, Y size B etc.

Existing collections have been collected for a variety of specific reasons, and for very specific domains, intended to capture regularities of some  particular communication situation,  such as e.g. the ATIS corpus of air travel information requests \citep{hemphill1990atis}, meeting recordings \citep{garofolo2004nist}, telephone conversations \citep{canavan1997callhome} and broadcast news \citep{garofolo2004rich}. 

Further, many existing corpora present relatively clean audio scenarios, and from a limited selection of domains, often with a single speaker reading from a prepared text, such as the original TIMIT collection \citep{garofolo1990timit} or the many available broadcast news corpora, which have been used as data sets for speech retrieval experiments in both TREC \citep{garofolo2000trec} and CLEF \citep{federico2003clef}. These more formal settings or samples of formal content are useful for the study of acoustic qualities of human speech but represent a more idealised scenario than many practical audio processing tasks of interest today.

There are some collections of more naturally occurring conversational material such as the CALLHOME corpus  \citep{canavan1997callhome}, the Santa Barbara Corpus of Spoken American English \citep{dubois2005} or the TED talks corpus \citep{hasebe2015design}. While some of the content in such collections share characteristics with podcast material, podcasts' combination of unscripted and spontaneously organised discourse in a conversational setting, with turntaking, interviews, stretches of monologue, argumentation, and the inclusion of other audio material including non-speech segments is not yet represented in any collection of spoken language available with transcripts for research purposes.

%%%% HERE A LIST OF PREVIOUSLY KNOWN PODCAST DATA SETS FOUND BY AASISH. JUSSI WILL GO THRU THEM TO DETERMINE WHICH DESERVE MENTION.
% list of links: https://www.kaggle.com/listennotes/all-podcast-episodes-published-in-december-2017
% commercially purchaseable data set: https://datarade.ai/data-providers/picasso/datasets/metadata-of-all-public-podcasts
% https://data.world/brandon-telle/podcasts-dataset
% no transcripts: https://sites.google.com/site/nhutnguyensite/home/audio-podcast-dataset
% unclear what this is: https://github.com/odenizgiz/Podcasts-Data
% partially transcribed: no transcripts: https://www.listennotes.com/datasets/
% 3000 episodes: http://damiano.github.io/podcastsummaries/

This dataset represents the first large scale corpus of fully transcribed podcast data that we are aware of. Other transcribed podcast datasets of interest have only partial transcripts such as the one collated by  \citet{yang2019podcast}, who released a dataset of up to the first ten minutes of \textasciitilde 88K podcast episodes, with transcripts for \textasciitilde 46K of these, or are orders of magnitude smaller such as the Podcast Summaries data set of some 3,000 podcast episodes \citep{spina2017extracting}.

%Summarization has been studied both in the single-document and multi-document setting. Our task differs from the 2018 TREC real-time summarization track in that it is single-document summary. Each document to be summarized (a podcast episode) is very long (up to two hours of speech) and is represented as audio or the output of an automatic speech rocognizer. In addition, we supply an extremely large training corpus, in that every podcast episode in the 100,000 corpus comes with a pre-existing text summary. Finally the audio files are of highly variable quality, so ASR transcripts will very in fidelity. 

%The named-person utterance detection task is related to MediaEval’s Person Discovery Task from 2015 and 2016, which required participants to automatically discover people speaking in a raw TV broadcast, and tag them [2]. It differs in that we have only audio and no video to aid in the identification. In addition, TV broadcasts are generally very well produced with good sound quality, while podcasts can vary greatly in acoustic cquality and professionalism of the speakers, as well as including more varied dialects,. 

\section{Data}

\begin{figure}
\centering
\includegraphics[width=0.5\textwidth]{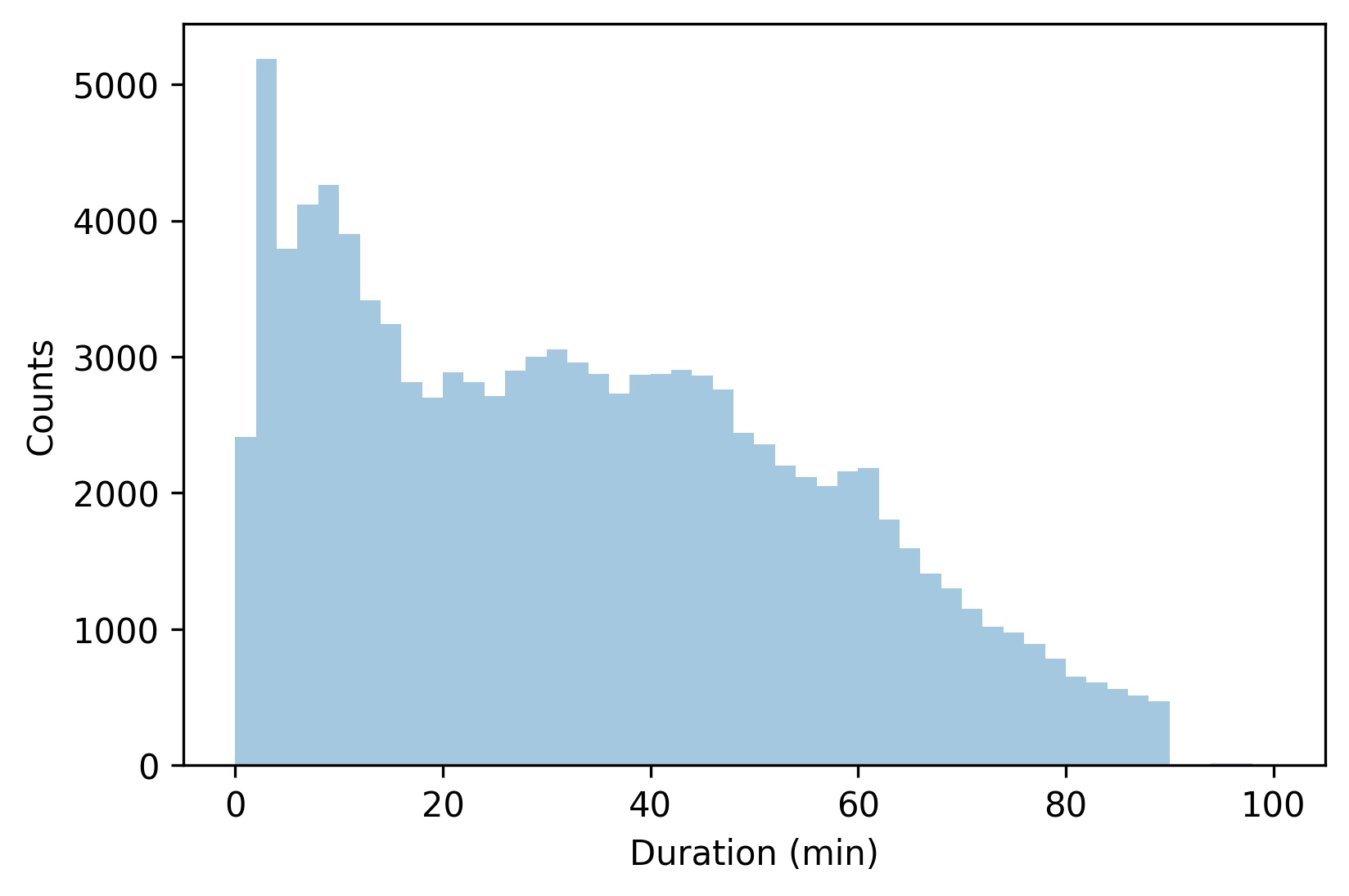}
\caption{Distribution of episode duration. We eliminated most podcasts longer than 90 minutes from the collection. Out of the 100K+ episodes, there were 125 that exceeded 90 minutes in length (not shown in graph).}
\label{fig:durations}
\end{figure}

This corpus consists of 100,000 podcast episodes, consisting of 50,000 hours of audio and accompanying transcripts.
These episodes are randomly sampled across a wide range of regions, domains, and audio quality. The episodes vary in length and include some very short trailers to advertise other content. Table~\ref{tab:my_label} gives descriptive statistics on word counts of podcast episodes in the dataset; Figure~\ref{fig:durations} gives the distribution over the lengths of the episodes in the dataset.

\begin{table}
    \centering
    \begin{tabular}{c|rrr}
                & min & average & max \\
    \hline
    minutes     &  $<$1 & 31.6 & 305.0\\
    words       &  11 & 5,728 & 43,504 \\
    \end{tabular}
    \caption{Descriptive statistics}
    \label{tab:my_label}
\end{table}

For each episode, we also supply the following metadata: show URI, episode URI, show name, episode name, show description, episode description, publisher, language, RSS link, and duration. In addition, we provide the full RSS header for each show, from which additional metadata can be extracted if desired.

\subsection{Episode Selection}

To select the episodes for inclusion in this dataset, we randomly sampled episodes published between January 1, 2019 and March 1, 2020, after filtering for several criteria.

For this dataset, we restricted episodes by language.
While in future versions of this dataset we would like to provide multilingual data, in the current version we limit it to English. 
In order to do this, we first restricted the set of episodes using metadata tags as provided by the creator specifying the language of the show to be English.
However, these tags can be inaccurate, so as an additional step, we run language identification\footnote{\url{https://pypi.org/project/langid/}} on the podcast description. 
We exclude the podcast title from the language identification step, as we found that titles are often very brief or named entities that yield low confidence predictions, whereas the descriptions are more frequently longer, grammatical utterances.
Since we recognize that many podcasts will have multilingual content and we wish to include this, we prefer to err on the side of precision when identifying non-English content, and therefore we only exclude those episodes that the identification tool positively identifies as being non-English.

We also did some filtering based on episode length.
For the professionally produced podcasts (those from Spotify Studios, Parcast, and Gimlet), we did not limit by length, since these were distributed towards shorter lengths, with only 125 episodes exceeding 90 minutes and a maximum length of 305 minutes.
However, for the non-professionally published episodes (those from Anchor), we restrict these in length to 90 minutes, as a heuristic quality filter, since excessively long amateur podcasts were more frequently subject to noise.

We further limited the dataset to include only those shows that had been streamed above a certain threshold number of times in the first month since their release. 
This heuristic provided us with a naive yet effective quality filter, removing defective or noisy material.

We also removed episodes that belonged to shows that averaged less than 50\% speech overall.
To do this, we used a classifier that automatically detects whether each one-second snippet of audio in the episode is noise, music, or speech.

Once these conditions were fulfilled, we then randomly sampled 100,000 episodes from the entire set.

\subsection{Transcription}

We generate the text transcripts automatically using Google Cloud Platform's Cloud Speech-to-Text API\footnote{\url{https://cloud.google.com/speech-to-text/docs/video-model}} (GCP-ASR), which provides word-level timestamps for each word. In addition, we include other inferred information it supplies: speaker diarization, casing, and punctuation insertion.

\begin{figure*}
\centering
\includegraphics[width=\textwidth]{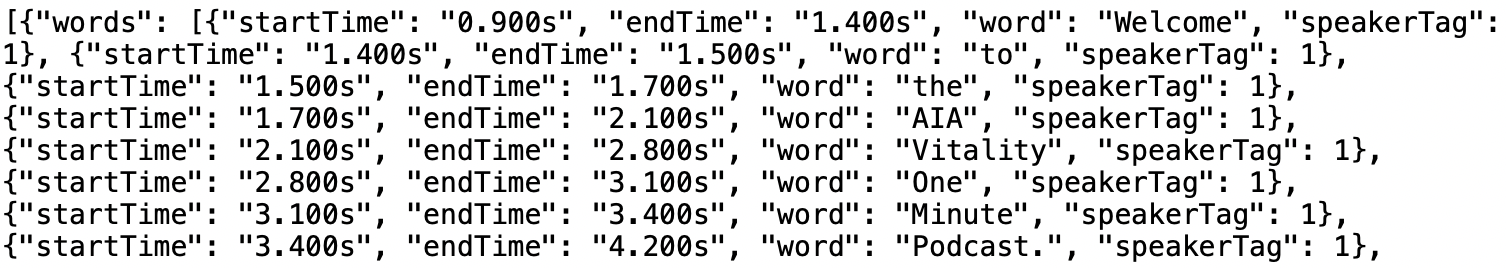}
\caption{transcript snippet}
\label{fig:transcript-example}
\end{figure*}

On a set of 1,600 English episodes, we conducted manual evaluation of the quality of the transcripts. 
We selected the episodes for the evaluation set to include a variety of domains, number of speakers, regional accents, level of formality, and audio quality. 
On our heterogeneous test set,  GCP-ASR showed robustness across the differences in the audio data quality and regional accents, with a word error rate of 18.1\% and an accuracy for named entity recognition of 81.8\%.
Figure~\ref{fig:transcript-example} shows an example snippet from a transcript.

% \begin{verbatim}
    
% [{"words": [{"startTime": "0.900s", "endTime": "1.400s", "word": "Welcome", "speakerTag": 1}, {"startTime": "1.400s", "endTime": "1.500s", "word": "to", "speakerTag": 1}, {"startTime": "1.500s", "endTime": "1.700s", "word": "the", "speakerTag": 1}, {"startTime": "1.700s", "endTime": "2.100s", "word": "AIA", "speakerTag": 1}, {"startTime": "2.100s", "endTime": "2.800s", "word": "Vitality", "speakerTag": 1}, {"startTime": "2.800s", "endTime": "3.100s", "word": "One", "speakerTag": 1}, {"startTime": "3.100s", "endTime": "3.400s", "word": "Minute", "speakerTag": 1}, {"startTime": "3.400s", "endTime": "4.200s", "word": "Podcast.", "speakerTag": 1}, 
% \end{verbatim}

\subsection{RSS Headers}

For each podcast episode, we provide the publicly available show-level RSS header {\sc HTML} file snapshot at the time of assembling these data.
These headers contain a wide variety of show-level and episode-level metadata, including the channel, title, description, author, link, copyright, language, image.
We advise that these RSS headers change over time, and are subject to noise.
For example, RSS headers may contain inaccurate episode-level information, such as links which are outdated.
Thus, while the transcripts we provide match the audio files in this dataset, they may not match pointers in the RSS feed.

\begin{figure*}
\centering
\includegraphics[width=\textwidth]{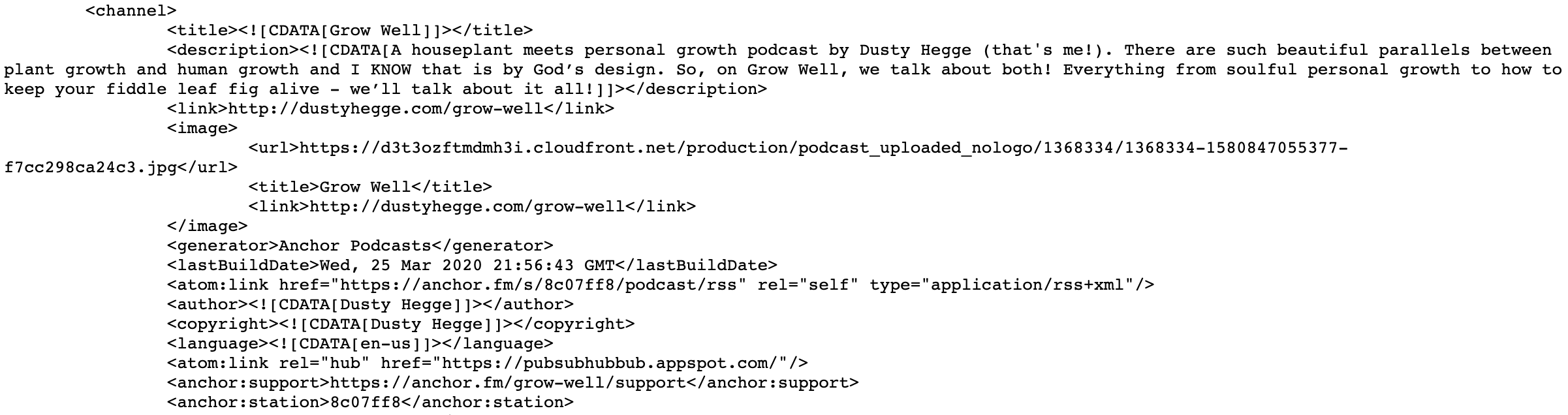}
\caption{RSS header snippet}
\label{fig:rss-example}
\end{figure*}
% here's the RSS html I grabbed this from: https://anchor.fm/s/8c07ff8/podcast/rss

Figure~\ref{fig:rss-example} shows an example of a podcast episode’s metadata from an excerpt from its RSS feed, which includes a pointer to the audio. 

\section{Conclusion}

We have presented the Spotify Podcast Dataset, the first large-scale dataset of podcasts with transcripts.
We hope that this will encourage researchers to explore the podcast medium in speech and text processing tasks, such as information retrieval and summarization.
\bibliographystyle{plainnat} 
\bibliography{main}

\begin{thebibliography}{12}
\providecommand{\natexlab}[1]{#1}
\providecommand{\url}[1]{\texttt{#1}}
\expandafter\ifx\csname urlstyle\endcsname\relax
  \providecommand{\doi}[1]{doi: #1}\else
  \providecommand{\doi}{doi: \begingroup \urlstyle{rm}\Url}\fi

\bibitem[Bois and Englebretson(2005)]{dubois2005}
John W.~Du Bois and Robert Englebretson.
\newblock Santa barbara corpus of spoken american english.
\newblock \emph{Linguistic Data Consortium}, 2005.
\newblock URL \url{https://catalog.ldc.upenn.edu/LDC2005S25}.

\bibitem[Canavan et~al.(1997)Canavan, Graff, and
  Zipperlen]{canavan1997callhome}
Alexandra Canavan, David Graff, and George Zipperlen.
\newblock Callhome american english speech.
\newblock \emph{Linguistic Data Consortium}, 1997.

\bibitem[Clifton et~al.(2020)Clifton, Reddy, Yu, Pappu, Rezapour, Bonab,
  Eskevich, Jones, Karlgren, Carterette, and Jones]{clifton-etal-2020-100000}
Ann Clifton, Sravana Reddy, Yongze Yu, Aasish Pappu, Rezvaneh Rezapour, Hamed
  Bonab, Maria Eskevich, Gareth Jones, Jussi Karlgren, Ben Carterette, and
  Rosie Jones.
\newblock 100,000 podcasts: A spoken {E}nglish document corpus.
\newblock In \emph{Proceedings of the 28th International Conference on
  Computational Linguistics}, pages 5903--5917, Barcelona, Spain (Online),
  December 2020. International Committee on Computational Linguistics.
\newblock URL \url{https://www.aclweb.org/anthology/2020.coling-main.519}.

\bibitem[Federico and Jones(2003)]{federico2003clef}
Marcello Federico and Gareth~JF Jones.
\newblock The clef 2003 cross-language spoken document retrieval track.
\newblock In \emph{Workshop of the Cross-Language Evaluation Forum for European
  Languages}, pages 646--652. Springer, 2003.

\bibitem[Garofolo et~al.(1990)Garofolo, Lamel, Fisher, Fiscus, Pallett,
  Dahlgren, and Zue]{garofolo1990timit}
John~S Garofolo, Lori~F Lamel, William~M Fisher, Jonathan~G Fiscus, David~S
  Pallett, Nancy~L Dahlgren, and Victor Zue.
\newblock Timit acoustic-phonetic continuous speech corpus.
\newblock \emph{Linguistic Data Consortium}, 1990.
\newblock URL \url{https://catalog.ldc.upenn.edu/LDC93S1}.

\bibitem[Garofolo et~al.(2000)Garofolo, Auzanne, and
  Voorhees]{garofolo2000trec}
John~S Garofolo, Cedric~GP Auzanne, and Ellen~M Voorhees.
\newblock The trec spoken document retrieval track: A success story.
\newblock \emph{NIST SPECIAL PUBLICATION}, 500\penalty0 (246):\penalty0
  107--130, 2000.

\bibitem[Garofolo et~al.(2004{\natexlab{a}})Garofolo, Fiscus, and
  Le]{garofolo2004rich}
John~S. Garofolo, Jonathan Fiscus, and Audrey Le.
\newblock Rich transcription broadcast news and conversational telephone speech
  . web download.
\newblock \emph{Linguistic Data Consortium}, 2004{\natexlab{a}}.
\newblock URL \url{https://catalog.ldc.upenn.edu/LDC2004S11}.

\bibitem[Garofolo et~al.(2004{\natexlab{b}})Garofolo, Laprun, Michel, Stanford,
  and Tabassi]{garofolo2004nist}
John~S Garofolo, Christophe Laprun, Martial Michel, Vincent~M Stanford, and
  Elham Tabassi.
\newblock The {NIST} meeting room pilot corpus.
\newblock In \emph{Proceedings of the 4th Conference on Language Resources and
  Evaluation (LREC)}. European Language Resources Association,
  2004{\natexlab{b}}.

\bibitem[Hasebe(2015)]{hasebe2015design}
Yoichiro Hasebe.
\newblock Design and implementation of an online corpus of presentation
  transcripts of ted talks.
\newblock \emph{Procedia-Social and Behavioral Sciences}, 198:\penalty0
  174--182, 2015.

\bibitem[Hemphill et~al.(1990)Hemphill, Godfrey, and
  Doddington]{hemphill1990atis}
Charles~T Hemphill, John~J Godfrey, and George~R Doddington.
\newblock The atis spoken language systems pilot corpus.
\newblock In \emph{Speech and Natural Language: Proceedings of a Workshop Held
  at Hidden Valley, Pennsylvania, June 24-27, 1990}, 1990.

\bibitem[Spina et~al.(2017)Spina, Trippas, Cavedon, and
  Sanderson]{spina2017extracting}
Damiano Spina, Johanne~R. Trippas, Lawrence Cavedon, and Mark Sanderson.
\newblock Extracting audio summaries to support effective spoken document
  search.
\newblock \emph{Journal of the Association for Information Science and
  Technology}, 68\penalty0 (9):\penalty0 2101--2115, 2017.
\newblock ISSN 2330-1643.
\newblock \doi{10.1002/asi.23831}.
\newblock URL \url{http://dx.doi.org/10.1002/asi.23831}.

\bibitem[Yang et~al.(2019)Yang, Wang, Dunne, Sobolev, Naaman, and
  Estrin]{yang2019podcast}
Longqi Yang, Yu~Wang, Drew Dunne, Michael Sobolev, Mor Naaman, and Deborah
  Estrin.
\newblock More than just words: Modeling non-textual characteristics of
  podcasts.
\newblock In \emph{Proceedings of the Twelfth ACM International Conference on
  Web Search and Data Mining}. ACM, 2019.

\end{thebibliography}
\end{document}